\documentclass[letterpaper, 10 pt, conference]{ieeeconf}  

\IEEEoverridecommandlockouts                              

\usepackage{graphics} 
\usepackage{epsfig} 
\usepackage{mathptmx} 
\usepackage{times} 
\usepackage{amsmath} 
\usepackage{amssymb}  
\usepackage{float} 
\usepackage{soul} 
\usepackage{xcolor}
\definecolor{softblue}{RGB}{200, 220, 255}
\sethlcolor{softblue}
\usepackage{subcaption}
\usepackage{tikz}
\usepackage{lipsum}
\usepackage{todonotes}

\usetikzlibrary{arrows.meta, positioning, calc, fit}
\title{\LARGE \bf
Contact-Aware Impedance Controller for Robot-Assisted Ultrasound Imaging 
}

\author{MD Miraj Arefin$^{1}$ and Mehmet Efe Tiryaki$^{1}$
\thanks{*This work was supported by the Scientific and Technological Research Council of Türkiye (TÜBİTAK) under Grant Number 124C815}
\thanks{$^{1}$ Center for Robotics and AI (ROMER),
        Middle East Technical University, Ankara, Turkey}
\thanks{*Corresponding Author's email: etiryaki@metu.edu.tr}
}

\begin{document}
\bstctlcite{IEEEexample:BSTcontrol}

\maketitle
\thispagestyle{empty}
\pagestyle{empty}

\begin{abstract}
Safe robot-assisted ultrasound imaging requires a reliable controller able to detect and localize probe--tissue interaction. In this paper, we present a B-mode ultrasound image-based contact perception method and a contact-aware impedance controller for robotic ultrasound imaging. The proposed method detects acoustic contact independently of force measurements, enabling contact-conditioned force/torque taring to reduce residual wrench bias. During contact, the method continuously estimates the effective contact location along the curved probe surface and uses it to update the controller interaction frame, enabling visual servoing of the physical probe--tissue contact point during imaging. Experiments on an agar phantom demonstrated a contact-localization RMSE of $\mathbf{1.46 \pm 0.14}$~mm over probe roll angles from $\mathbf{-15^\circ}$ to $\mathbf{15^\circ}$. During static rolling, the proposed controller maintained task-space tracking accuracy comparable to a conventional fixed-frame impedance controller while reducing the maximum compressive interaction force from $\mathbf{31.56}$~N to $\mathbf{20.09}$~N, corresponding to a $\mathbf{36.3\%}$ reduction. These results demonstrate the potential of ultrasound images as direct contact feedback for safe and accurate robot-assisted ultrasound imaging.
\end{abstract}

\section{INTRODUCTION}

Ultrasound (US) is widely used for diagnostic and interventional imaging because it provides real-time, non-ionizing, and comparatively low-cost visualization of internal anatomy \cite{izadifar2017mechanical}. Despite these advantages, conventional US examination remains highly dependent on the operator. Image quality is affected not only by the selected imaging location, but also by probe orientation, contact force, and the resulting acoustic coupling with the tissue \cite{chatelain2015optimization,chatelain2017confidence}. Consequently, experienced sonographers continuously adjust both probe pose and applied pressure while interpreting the acquired images. This dependence on manual manipulation introduces inter- and intra-operator variability and makes prolonged examinations physically demanding \cite{vanderpool1993prevalence,harrison2015work}. Robot-assisted ultrasound (RUS) systems  offer a means to standardize probe motion and contact conditions, improve acquisition repeatability, enable accurately localized volumetric scans, and reduce the physical burden on clinicians \cite{welleweerd2020automated,ma2022see}. Recent demonstrations in thyroid, lung, vascular, and breast imaging further illustrate the progression toward increasingly autonomous US acquisition \cite{su2024fully,lei2025toward,lee2024combining,zhang2025smooth}. Achieving such autonomy requires reliable perception of probe--tissue interaction. For curved probes, the effective contact location can shift during reorientation, making B-mode ultrasound a useful feedback modality for contact-aware visual servoing.

\begin{figure}[t]
\includegraphics[width=\linewidth]{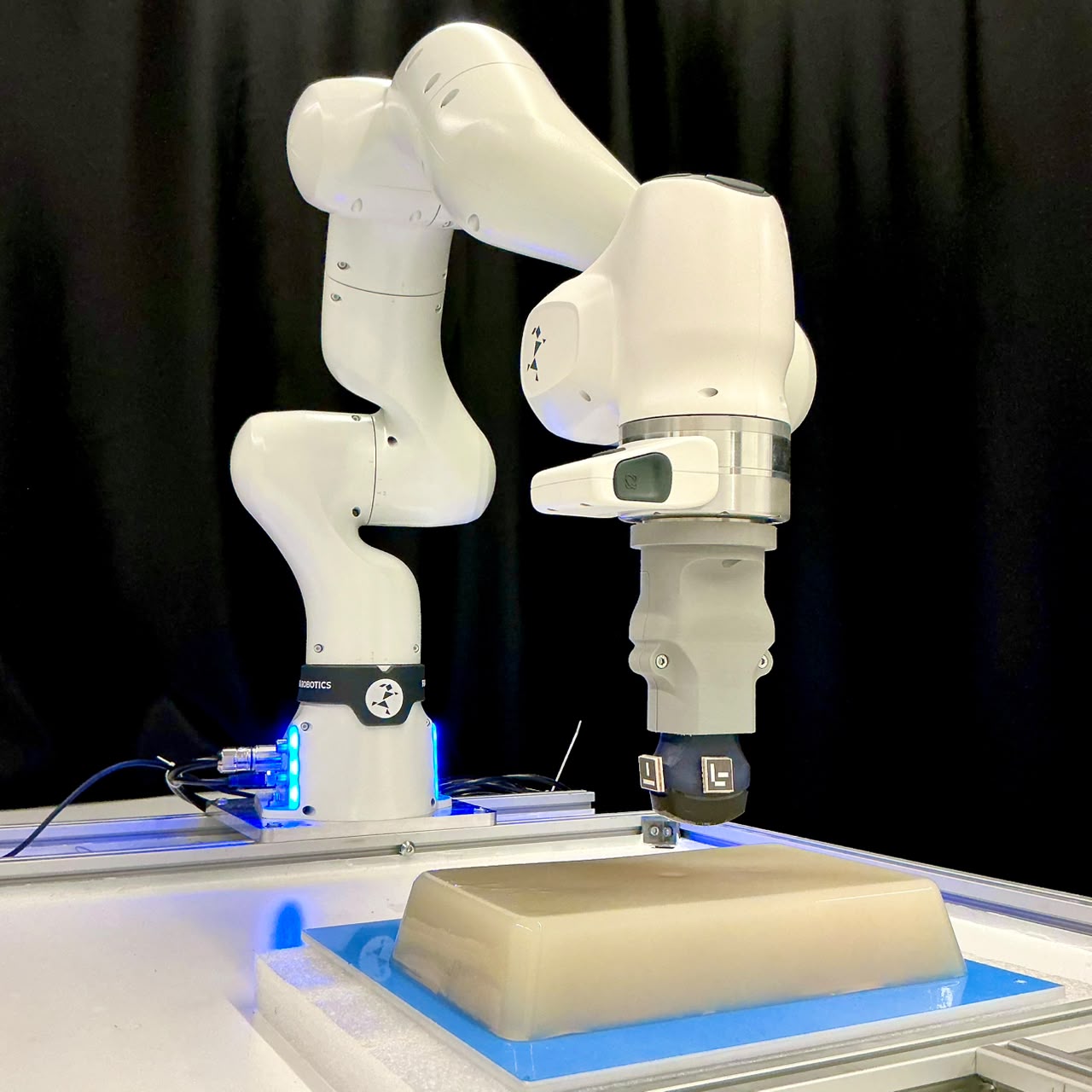}
\caption{The experimental setup for RUS system.}
\label{fig:figure_1}
\vspace{-10pt}
\end{figure}

\subsection{Related Work}

Recent studies on autonomous RUS focus on integrating perception, planning, and compliant control \cite{jiang2023robotic,guo2026technical}. A central control problem is maintaining reliable acoustic coupling while limiting tissue loading. Accordingly, admittance \cite{carriere2019admittance, jiang2024force}, impedance \cite{fu2024optimization}, and explicit force-control \cite{piedra2026explicit} strategies have been developed to regulate probe--tissue interaction. These approaches can maintain a prescribed contact force and improve robustness to uncertain or moving surfaces. However, force regulation alone does not determine where the resultant interaction acts on a curved transducer, which becomes relevant when probe reorientation causes the effective interaction region to migrate along the probe surface.

The spatial relationship between the probe and tissue has therefore been addressed using geometric and mechanical sensing. Active optical and RGB-D sensing have been used to estimate the local tissue surface and surface normal for probe approach and orientation control \cite{ma2022see,zhetpissov2025asee2,lee2024combining}. Other approaches infer interaction geometry from force measurements. Jiang et al. estimated the local surface normal from the force response during controlled probe rotations \cite{jiang2021forcebased}, while Cao et al. analytically recovered the mechanical contact location from force/torque measurements and instrument geometry and incorporated this estimate into RUS control \cite{cao2023intrinsic}. Xiao et al. further combined force and position measurements with ultrasound-confidence information to estimate the environmental contact location and stiffness \cite{xiao2025variable}. These methods provide valuable geometric or mechanical information about the probe--tissue interface, but their spatial contact estimate depends on external geometric sensing and/or force measurements.

\begin{figure}[t]
\vspace*{5pt} 
\includegraphics[width=\linewidth]{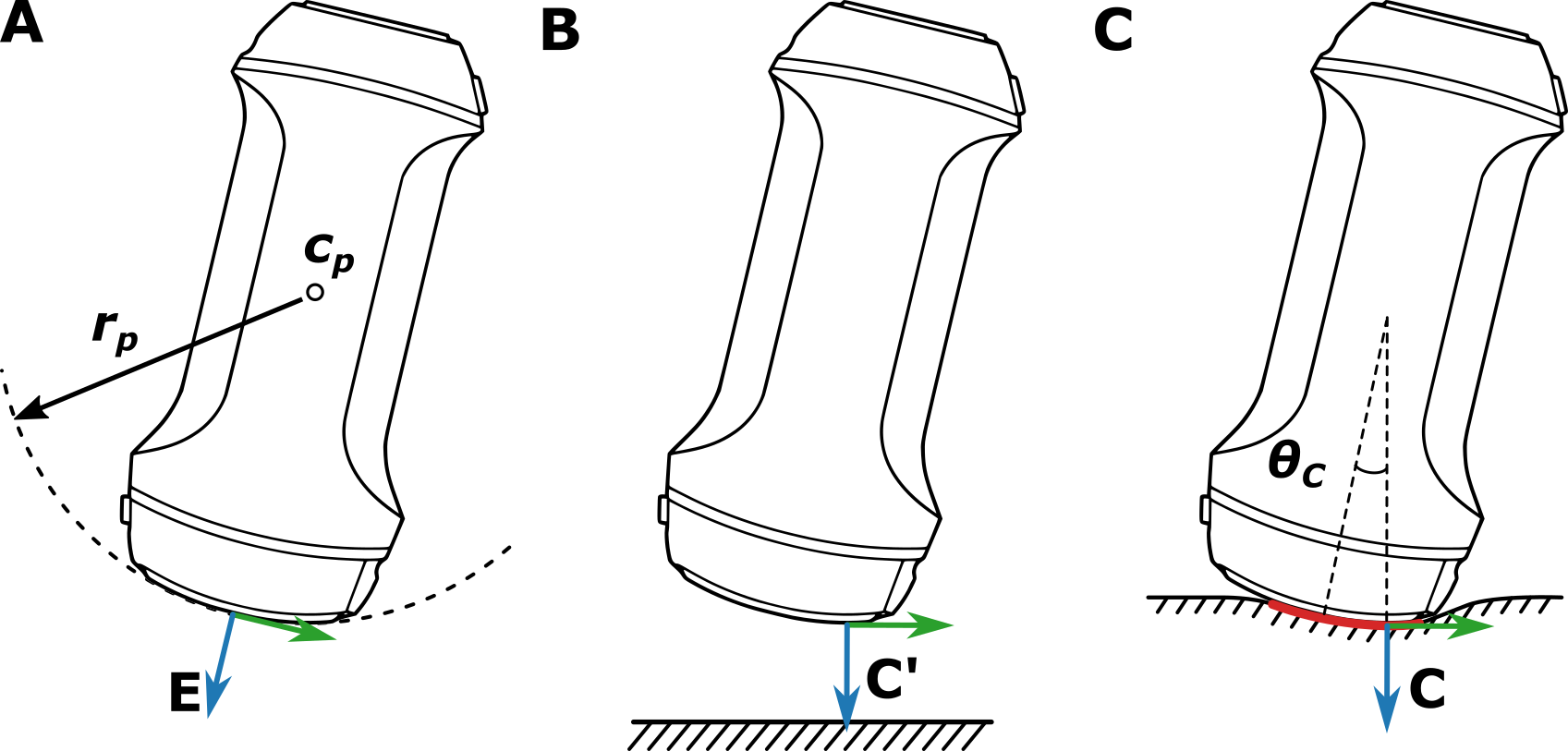}
\caption{Contact Point Geometry. A) The probe radius and fixed control frame. The contact frame during B) the non-contact phase and C) the in-contact phase. The red curve is the US probe contact region. }
\label{fig:cp_geom}
\end{figure}

Ultrasound images provide a complementary feedback modality directly related to acoustic coupling. Intensity- and confidence-based visual servoing have been used to regulate probe pose and optimize image quality \cite{chatelain2017confidence,nadeau2011intensity,jiang2020automatic}. In particular, Chatelain et al. combined confidence-based orientation control with force regulation in a probe contact frame, whose relative transformation was experimentally calibrated \cite{chatelain2015optimization}. Welleweerd et al. used the mean and barycenter of an ultrasound confidence map to identify incomplete probe contact and command axial and in-plane corrections during robotic breast scanning \cite{welleweerd2020automated}. Akbari et al. further derived image-based contact information by comparing the B-mode image with a no-contact reference and used image-quality features to adjust the applied force  \cite{akbari2021robotic}. Jiang et al. used an ultrasound confidence map and its weighted barycenter to detect incomplete contact and correct probe orientation after repositioning \cite{jiang2022repositioning}. Similarly, Huang et al. showed that constant force does not guarantee complete probe--skin contact and used B-mode grayscale asymmetry and anatomical features to guide probe orientation and autonomous carotid scanning \cite{huang2024carotid}. These studies establish that B-mode ultrasound can provide useful information about acoustic coupling; however, the image feedback is primarily converted into image-quality, pose, or force-control objectives rather than a continuous spatial estimate of the effective acoustic contact location on the probe surface. 

This distinction becomes important for curved probes during rolling. As the probe rotates, the effective probe--tissue interaction region can migrate along the curved face, such that the kinematics of the interaction location no longer coincide with those of a fixed end-effector or
pre-calibrated contact point. Consequently, regulating interaction about a fixed control frame can produce unintended motion or force at the effective contact location during probe reorientation. To the best of our knowledge, prior RUS work has not jointly used B-mode ultrasound to detect acoustic
contact independently of the force/torque signal, continuously localize the effective acoustic contact location along a curved probe, and update the task-space interaction frame according to this moving image-derived estimate. This gap motivates the contact-aware formulation developed in this work.

\subsection{Contribution}
We propose a B-mode US-based contact perception framework and a contact-aware impedance controller for robot-assisted US imaging, evaluated using the experimental RUS system shown in Fig.~\ref{fig:figure_1}. The framework detects acoustic contact independently of force sensing and estimates the effective contact location along the curved probe surface. The controller uses this estimate to regulate contact-point position and end-effector orientation as contact moves during probe rolling. In this way, the proposed framework implements US image-based visual servoing, where B-mode image information provides online spatial feedback for robot–tissue interaction control. A geometry-based potential contact point provides the control reference before contact.

The main contributions are:
\begin{itemize}
    \item \textbf{US-based contact perception:}
    A B-mode intensity-based method for detecting probe--tissue contact and continuously localizing the effective contact point, without relying on force measurements.

    \item \textbf{Contact-aware interaction control:}
    An impedance controller that updates its controlled position and wrench application point using the estimated contact location, enabling compliant interaction during probe rolling.
\end{itemize}

\section{Contact Detection and Localization}
During acoustic contact, the probe couples with tissue over a finite contact region. We refer to the mechanical center of pressure of this region as the \emph{instantaneous contact point}, where the resultant normal contact force acts. Friction can complicate its localization from joint-torque-based wrench estimates. We therefore hypothesize that we can localize this contact point from B-mode US images.

\subsection{Contact Point Geometry}
\label{subsec:Contact Point Geometry}
As the curved probe rolls over the tissue, the contact
location moves along the probe surface. We represent this
location by a contact frame whose position and orientation
are updated from the estimated contact geometry
(Fig.~\ref{fig:cp_geom}).

We denote the robot base, end-effector, and instantaneous
contact frames by $\mathbf{O}$, $\mathbf{E}$, and
$\mathbf{C}$, respectively. Left superscripts indicate
the frame in which a quantity is expressed. The
end-effector position and orientation in the base frame
are ${}^{O}\mathbf{p}_{E}\in\mathbb{R}^{3}$ and
${}^{O}\mathbf{R}_{E}\in SO(3)$. As the probe rolls over the tissue, the contact point
moves along its curved surface in the $YZ$ plane of
$\mathbf{E}$. The angle $\theta_C$ specifies this point's
location on the arc, measured from the end-effector's
positive $z$-axis about its $x$-axis. We orient the contact
frame so that its $z$-axis points radially toward the
contact point:
\begin{equation}
    {}^{O}\mathbf{R}_{C}
    =
    {}^{O}\mathbf{R}_{E}\mathbf{R}_{x}(\theta_C),
    \label{eq:contact_rotation}
\end{equation}
where $\mathbf{R}_{x}(\theta_C)$ denotes rotation about
the local $x$-axis. The contact and end-effector axes
are aligned at $\theta_C=0$.

We define the probe's center of curvature in the
end-effector frame as ${}^{E}\mathbf{c}_p$ and its
radius of curvature as $r_p$. The contact-point
position in the base frame is
\begin{equation}
    {}^{O}\mathbf{p}_{C}
    =
    {}^{O}\mathbf{p}_{E}
    +
    {}^{O}\mathbf{R}_{E}
    \left(
        {}^{E}\mathbf{c}_{p}
        +
        r_p\,\mathbf{R}_x(\theta_C)\mathbf{e}_3
    \right),
    \label{eq:contact_position}
\end{equation}
where $\mathbf{e}_3=[0,0,1]^T$. Thus, knowing the
contact angle $\theta_C$ allows us to estimate the
contact-point position.

\subsection{Contact State Detection}
We defined the raw B-mode ultrasound image as a 2D matrix $\mathbf{I} \in \mathbb{R}^{M \times N}$, where $M=512$ is the number of rows along the vertical image direction  and $N=128$ is the number of columns along the horizontal image direction. The intensity of the pixel at axial row $m$ and
lateral column $j$ is denoted by $I_{m,j}$, with $m\in\{0,\ldots,M-1\}$ and $j\in\{0,\ldots,N-1\}$. The two-dimensional image is projected to a 1D column-intensity profile by summing the pixel intensities along the image rows:
\begin{equation}
I_c(j)
=
\sum_{m=0}^{M-1} I_{m,j},
\label{eq:column_profile}
\end{equation}

\begin{figure}[t]
\vspace*{5pt} 
\includegraphics[width=\linewidth]{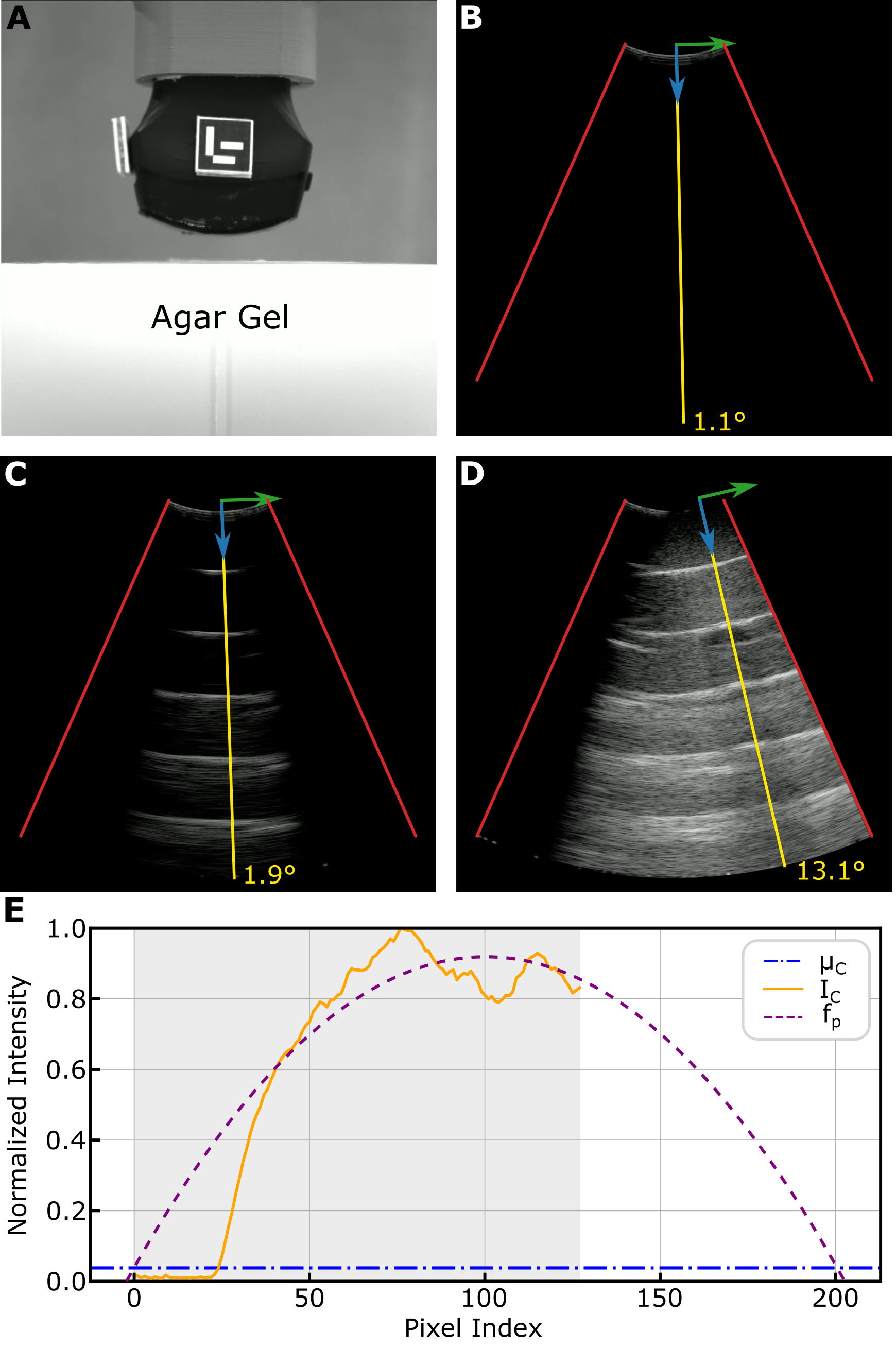}
\caption{US-based Contact Detection and Localization. A) The RUS probe and agar gel. Example US image B) during the non-contact phase. C)  when contact is detected, and D) the in-contact phase. E) The contact detection method. The gray shaded area shows the US imaging region.}
 \label{fig:parabola_fit}
\vspace{-20pt}
\end{figure}

Acoustic coupling between the probe and tissue produces a
substantial increase in the received B-mode echo intensity.
We exploit this change to detect the transition between the
\emph{non-contact} and \emph{in-contact} states independently
of the force/torque measurements. The peak intensity of the
column profile is defined as $A_c = \max_{j} I_c(j)$
and the contact state is determined as
\begin{equation}
S_c =
\begin{cases}
1, & A_c > \mu_c,\\[1mm]
0, & \text{otherwise},
\end{cases}
\label{eq:contact_state}
\end{equation}
where $\mu_c$ is an experimentally selected contact-detection
threshold. In the experiments, $\mu_c=2000$ was used for the
acquired B-mode images. An example of the image corresponding
to the transition into contact is shown in
Fig.~\ref{fig:parabola_fit}C.

\subsection{Contact Localization}
To provide a unified controller across contact states,
we estimate a potential contact point before contact and
use image-based localization during contact.

\emph{During the non-contact phase} $(S_c=0)$
(Fig.~\ref{fig:cp_geom}B), we estimate the tissue plane
near the probe. The probe's center of curvature in the
base frame is
${}^{O}\mathbf{c}_p
={}^{O}\mathbf{p}_E
+{}^{O}\mathbf{R}_E\,{}^{E}\mathbf{c}_p$.
We fit a local plane to the $k$ surface points nearest
to ${}^{O}\mathbf{c}_p$ ($k=10$), with all points
expressed in $\mathbf{O}$. The plane passes through
their centroid ${}^{O}\bar{\mathbf{s}}$, and its unit
normal ${}^{O}\mathbf{n}_s$ is obtained by principal
component analysis as the direction of least spatial
variance. Its sign is selected to point toward the
probe center.

Projecting the probe center onto this plane gives
\begin{equation}
{}^{O}\mathbf{p}_{\Pi}
=
{}^{O}\mathbf{c}_p
-
{}^{O}\mathbf{n}_s
\left({}^{O}\mathbf{n}_s\right)^{T}
\left({}^{O}\mathbf{c}_p-{}^{O}\bar{\mathbf{s}}\right),
\label{eq:surface_projection}
\end{equation}
where ${}^{O}\mathbf{p}_{\Pi}$ lies on the estimated
tissue plane. The direction from the probe center
toward this plane is
\begin{equation}
{}^{O}\hat{\mathbf{r}}_{P}
=
\frac{{}^{O}\mathbf{p}_{\Pi}-{}^{O}\mathbf{c}_p}
{\|{}^{O}\mathbf{p}_{\Pi}-{}^{O}\mathbf{c}_p\|}.
\label{eq:potential_radial_direction}
\end{equation}
We express this direction in the end-effector frame as
${}^{E}\hat{\mathbf{r}}_{P}
=({}^{O}\mathbf{R}_{E})^{T}\,{}^{O}\hat{\mathbf{r}}_{P}$,
then apply the cylindrical projection and normalization described in the Sec.~\ref{subsec:Contact Point Geometry} to select a point on the probe surface facing the tissue. This potential contact point defines the control point during approach; it does not indicate that physical contact has occurred. Once $S_c=1$, the controller instead uses the contact location estimated from the B-mode image.

\emph{During the in-contact phase} $(S_c=1)$
(Fig.~\ref{fig:cp_geom}C), the effective contact location is
estimated from the lateral distribution of B-mode intensity.
As the probe rolls over the tissue, the acoustically coupled
region migrates along the curved probe surface, producing a
corresponding displacement of the column-intensity distribution.

For localization, the raw profile in \eqref{eq:column_profile} is normalized by its peak value as $\bar{I}_c(j)=I_c(j)/A_c$, for $j\in{0,\ldots,N-1}$. Thus, $\bar{I}_c(j)\in[0,1]$. The center of the intensity distribution is estimated by fitting
a constrained downward-opening parabola. Using $u$ as the
continuous lateral image coordinate, the model is
\begin{equation}
f_p(u;a,b)
=
a
\left[
1-
\left(
\frac{u-b}{W}
\right)^2
\right],
\label{eq:parabolic_model}
\end{equation}
where $a\geq0$ is the fitted peak amplitude, $b$ is the
continuous lateral coordinate of the parabola vertex, and $W$
is a fixed width parameter expressed in image-column units.
The parabola reaches zero at $u=b\pm W$. In this work,
$W=N/1.25$ was selected empirically and kept constant for all
experiments, where $N$ is the number of lateral image columns.

To reduce the influence of low-intensity background regions,
only columns whose normalized intensity exceeds $20\%$ of the
profile maximum are included in the fit. The corresponding set
of valid column indices is
\begin{equation}
\mathcal{V}
=
\left\{
j\in\{0,\ldots,N-1\}
\;\middle|\;
\bar{I}_c(j)>0.2
\right\}.
\label{eq:valid_columns}
\end{equation}

The parabola parameters are estimated by solving
\begin{equation}
(a^\ast,b^\ast)
=
\underset{
a\geq0,\;
0\leq b\leq N-1
}{\arg\min}
\sum_{j\in\mathcal{V}}
\left[
\bar{I}_c(j)
-
f_p(j;a,b)
\right]^2 .
\label{eq:parabola_fit}
\end{equation}

The fitted vertex $b^\ast$ provides a continuous estimate of the
center of the acoustic contact distribution and therefore enables
sub-column localization. The estimated lateral coordinate is
subsequently mapped to the angular coordinate of the curved probe.
For a symmetric probe arc angle $\Psi_p$, the contact angle is
\begin{equation}
\theta_C
=
\left(
\frac{
b^\ast-\frac{N-1}{2}
}{
\frac{N-1}{2}
}
\right)
\frac{\Psi_p}{2},
\label{eq:image_to_contact_angle}
\end{equation}
where $\Psi_p$ denotes the angular span of the curved probe
surface represented in the B-mode image
($\Psi_p = 48^\circ$ for the probe used in this study). The estimated angle $\theta_C$ is then substituted into the probe geometry in Eqs.~\eqref{eq:contact_rotation} and \eqref{eq:contact_position} to update the contact-frame orientation ${}^{O}\mathbf{R}_C$ and contact-point position ${}^{O}\mathbf{p}_C$, respectively.

\section{Controller Design}


\subsection{Impedance Controller}

We formulate an impedance controller that regulates the
instantaneous contact-point position and the end-effector
orientation. Updating the controlled position as contact
moves along the curved probe allows the robot to roll
while maintaining the desired contact location.
Rotational feedback acts directly on the end-effector
orientation.

We define the desired contact-point position and
end-effector orientation in the base frame as
${}^{O}\mathbf{p}_{d}\in\mathbb{R}^{3}$ and
${}^{O}\mathbf{R}_{d}\in SO(3)$, respectively.
The translational and rotational errors are
\begin{equation}
    {}^{O}\tilde{\mathbf{x}}_{p}
    =
    {}^{O}\mathbf{p}_{d}
    -
    {}^{O}\mathbf{p}_{C},
    \label{eq:position_error}
\end{equation}
\begin{equation}
    {}^{O}\tilde{\mathbf{x}}_{R}
    =
    \left(
        \log\!\left(
            {}^{O}\mathbf{R}_{d}
            ({}^{O}\mathbf{R}_{E})^{T}
        \right)
    \right)^{\vee},
    \label{eq:ee_orientation_error}
\end{equation}
where $\log(\cdot)$ is the matrix logarithm and
$(\cdot)^{\vee}$ maps a skew-symmetric matrix to its
corresponding vector. The complete task-space pose
error is
\begin{equation}
    {}^{O}\tilde{\mathbf{x}}
    =
    \begin{bmatrix}
        {}^{O}\tilde{\mathbf{x}}_{p}\\
        {}^{O}\tilde{\mathbf{x}}_{R}
    \end{bmatrix}
    \in\mathbb{R}^{6}.
    \label{eq:task-space_error}
\end{equation}

\begin{figure}[t]
\vspace*{5pt} 
\centering
\includegraphics[width=\linewidth]{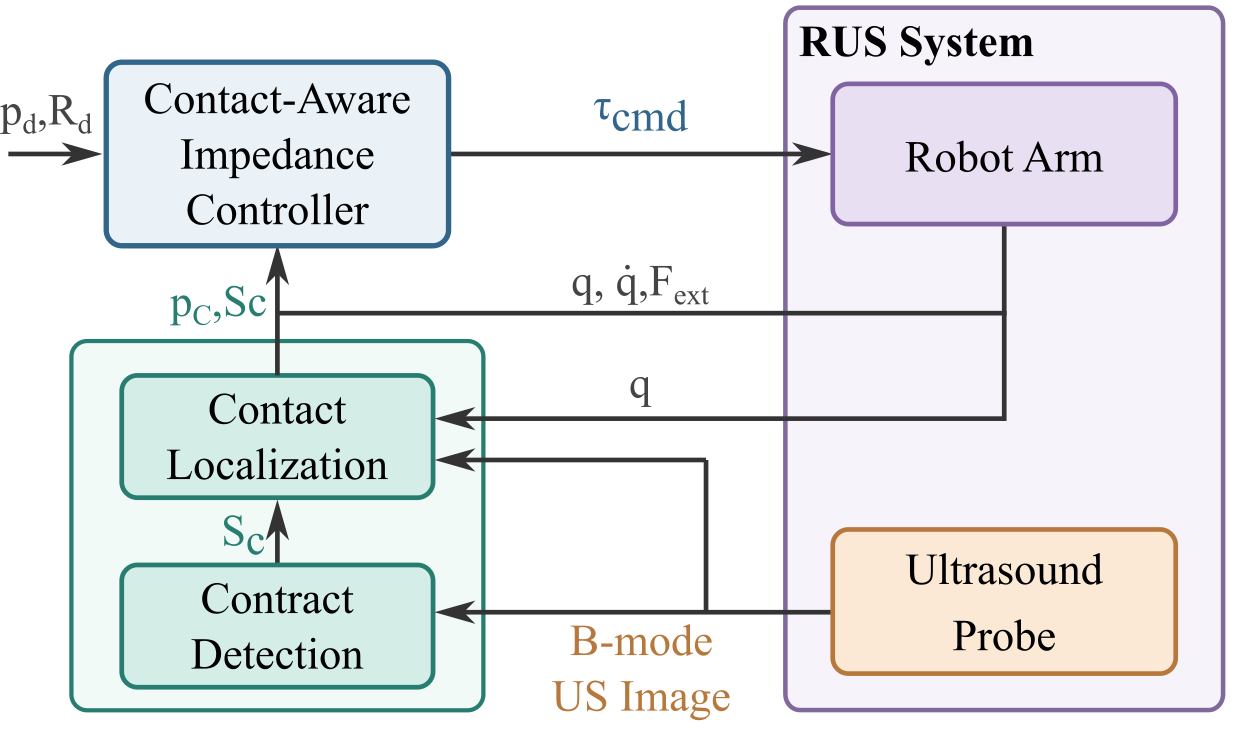}
\caption{Closed-loop architecture of the contact-aware impedance controller}
\label{fig:control_architecture}
\vspace{-10pt}
\end{figure}

We define the joint positions and velocities as
$\mathbf{q},\dot{\mathbf{q}}\in\mathbb{R}^{n}$, where
$n$ is the number of robot joints. The geometric
Jacobian ${}^{O}\mathbf{J}_{E}\in\mathbb{R}^{6\times n}$
maps joint velocities to the end-effector linear and
angular velocities, stacked in that order and expressed
in $\mathbf{O}$.

To apply the commanded wrench at the instantaneous
contact point, we shift this Jacobian from the
end-effector origin to the contact point. We define
the corresponding offset as
${}^{O}\mathbf{r}_{EC}
={}^{O}\mathbf{p}_{C}-{}^{O}\mathbf{p}_{E}$.
The shifted Jacobian is
\begin{equation}
    {}^{O}\mathbf{J}_{C}
    =
    \begin{bmatrix}
        \mathbf{I}_{3}
        &
        -[{}^{O}\mathbf{r}_{EC}]_{\times}\\
        \mathbf{0}_{3\times3}
        &
        \mathbf{I}_{3}
    \end{bmatrix}
    {}^{O}\mathbf{J}_{E},
    \label{eq:shifted_jacobian}
\end{equation}
where $[\cdot]_{\times}$ denotes the skew-symmetric
cross-product matrix. This Jacobian describes the
rigid-body velocity at the current contact location;
its angular component is the end-effector angular
velocity ${}^{O}\boldsymbol{\omega}_{E}$.

The estimated contact point also moves relative to
the probe as $\theta_C$ changes. Using the probe
geometry defined previously, its additional
translational velocity in the base frame is
\begin{equation}
    {}^{O}\mathbf{v}_{\theta}
    =
    {}^{O}\mathbf{R}_{E}
    \left[
        \left(\dot{\theta}_{C}\mathbf{e}_{1}\right)
        \times
        \left(r_p\mathbf{R}_{x}(\theta_C)\mathbf{e}_{3}\right)
    \right],
    \label{eq:dynamic_cp_velocity}
\end{equation}
where $\dot{\theta}_C$ is the estimated contact-angle
rate, $\mathbf{e}_{1}=[1,0,0]^T$, and
$\mathbf{e}_{3}=[0,0,1]^T$. The velocity vector
associated with the controlled position and
orientation is therefore
\begin{equation}
    {}^{O}\mathbf{v}_{\mathrm{task}}
    =
    \begin{bmatrix}
        {}^{O}\dot{\mathbf{p}}_{C}\\
        {}^{O}\boldsymbol{\omega}_{E}
    \end{bmatrix}
    =
    {}^{O}\mathbf{J}_{C}\dot{\mathbf{q}}
    +
    \begin{bmatrix}
        {}^{O}\mathbf{v}_{\theta}\\
        \mathbf{0}_{3}
    \end{bmatrix}.
    \label{eq:task_velocity}
\end{equation}
Contact migration contributes only to the translational
component because the controlled orientation is that
of the end-effector.

We define the desired contact-point velocity and
end-effector angular velocity in the base frame as
${}^{O}\mathbf{v}_{d}$ and
${}^{O}\boldsymbol{\omega}_{d}$, respectively.
The velocity error used for damping is
\begin{equation}
    {}^{O}\mathbf{e}_{v}
    =
    \begin{bmatrix}
        {}^{O}\mathbf{v}_{d}\\
        {}^{O}\boldsymbol{\omega}_{d}
    \end{bmatrix}
    -
    {}^{O}\mathbf{v}_{\mathrm{task}}.
    \label{eq:velocity_error}
\end{equation}

The external wrench at the end-effector is estimated
from the robot's joint-torque measurements. We express
its force and moment components in the base frame as
${}^{O}\mathbf{f}_{E}$ and ${}^{O}\mathbf{m}_{E}$.
Shifting the moment reference to the instantaneous
contact point gives
\begin{equation}
    {}^{O}\mathbf{F}_{\mathrm{ext}}
    =
    \begin{bmatrix}
        {}^{O}\mathbf{f}_{E}\\
        {}^{O}\mathbf{m}_{E}
        -
        {}^{O}\mathbf{r}_{EC}\times{}^{O}\mathbf{f}_{E}
    \end{bmatrix}.
    \label{eq:external_wrench}
\end{equation}

The commanded task-space wrench, expressed in the
base frame and referenced to the contact point, is
\begin{equation}
    {}^{O}\mathbf{F}_{\mathrm{cmd}}
    =
    {}^{O}\mathbf{K}_{d}\,{}^{O}\tilde{\mathbf{x}}
    +
    {}^{O}\mathbf{D}_{d}\,{}^{O}\mathbf{e}_{v}
    +
    {}^{O}\mathbf{F}_{\mathrm{ext}},
    \label{eq:wrench_command}
\end{equation}
where
${}^{O}\mathbf{K}_{d},{}^{O}\mathbf{D}_{d}
\in\mathbb{R}^{6\times6}$
are the task-space stiffness and damping matrices. The task-space stiffness matrix was selected as
$\mathbf{K}_{d} = \mathrm{diag}(1500,\,1500,\,1500,\,100,\,100,\,100)$ where the first three entries correspond to translational stiffness in $\mathrm{N/m}$ and the last three to rotational stiffness in $\mathrm{N\,m/rad}$. The damping matrix was selected according to the critical-damping condition $D_{d}=2\zeta\sqrt{\mathbf{K}_{d}\mathbf{M}_d}$. The ${}^{O}\mathbf{F}_{\mathrm{ext}}$ term applies feedback of the estimated
external wrench.

The task wrench is mapped to joint torques and
combined with the robot's gravity and dynamics
compensation. We denote these compensation torques
collectively by
$\boldsymbol{\tau}_{\mathrm{comp}}\in\mathbb{R}^{n}$.
The resulting joint-torque command is
\begin{equation}
    \boldsymbol{\tau}_{\mathrm{cmd}}
    =
    \left({}^{O}\mathbf{J}_{C}\right)^{T}
    {}^{O}\mathbf{F}_{\mathrm{cmd}}
    +
    \boldsymbol{\tau}_{\mathrm{comp}}.
    \label{eq:torque_command}
\end{equation}

\subsection{Adaptive Force Taring}
\label{subsec:taring}

Force/torque taring is required to remove residual sensor offsets caused by sensor bias, payload-model mismatch, and slow drift, which would otherwise appear as a false interaction wrench and degrade force regulation. In the proposed framework, taring is conditioned on the US-based contact state. During the non-contact phase ($S_c=0$), the raw six-axis wrench measurements are continuously stored in a fixed-length buffer. Upon a transition from non-contact to contact ($S_c:0\rightarrow1$), the most recent samples are discarded to avoid measurements potentially affected by the onset of contact, and the tare offset is computed as the mean of the remaining pre-contact measurements. A maximum of 1000 samples is stored which is collected at 1000 Hz, with the most recent 400 samples after contact is discarded, leaving up to 600 samples for tare estimation. After contact is lost, wrench samples are collected again following a short settling interval to prepare the tare estimate for the next contact event.

\section{Results}

\subsection{Experimental Setup}

The experimental platform is shown in Fig.~\ref{fig:figure_1}. A Franka Research 3 (Franka Robotics GmbH, Munich, Germany) robotic manipulator was equipped with a convex US probe (Konted C10RL Pro, Beijing Konted Medical Technology Co., Ltd., Beijing, China) and used for the contact and scanning experiments. The probe has a curved imaging surface with a radius of curvature of $50~\mathrm{mm}$ and an angular span of approximately $48^\circ$. B-mode US images were streamed over Wi-Fi at a resolution of $128\times512$ pixels and an acquisition rate of approximately $9~\mathrm{Hz}$. The images were processed online for contact-state detection and estimation of the instantaneous probe--phantom contact location.

All experiments were conducted on a flat agar phantom prepared with a $4\%$ (w/v) agar--agar concentration in water. Under static conditions, a probe indentation of $3~\mathrm{mm}$ below the undeformed phantom surface produced a measured normal force of approximately $15~\mathrm{N}$. We therefore used a nominal indentation depth of $3~\mathrm{mm}$ in all experiments, with $15~\mathrm{N}$ as the reference normal force. A six-axis force/torque sensor (Robotiq FT~300, Robotiq Inc., Quebec, Canada), operating at $100~\mathrm{Hz}$, was placed beneath the agar phantom to provide an independent measurement of the probe--phantom interaction force. All quantitative values were presented as means ± SD. 

\subsection{Contact Point Estimation Performance}
\label{subsec:CPEstPer}

We evaluated the contact-point assumption by holding the desired translational position constant while commanding sinusoidal probe rolling between $-15^\circ$ and $+15^\circ$. We compared the image-derived contact location with the probe's downward-facing point, which is expected to coincide with the mechanical center of pressure for the uniform, planar agar phantom under approximately symmetric loading. This comparison assesses whether the US-based estimate captures the expected mechanical contact location as it moves along the curved probe surface.

\begin{figure}[t]

\vspace*{5pt} 
\centering
\includegraphics[width=\linewidth]{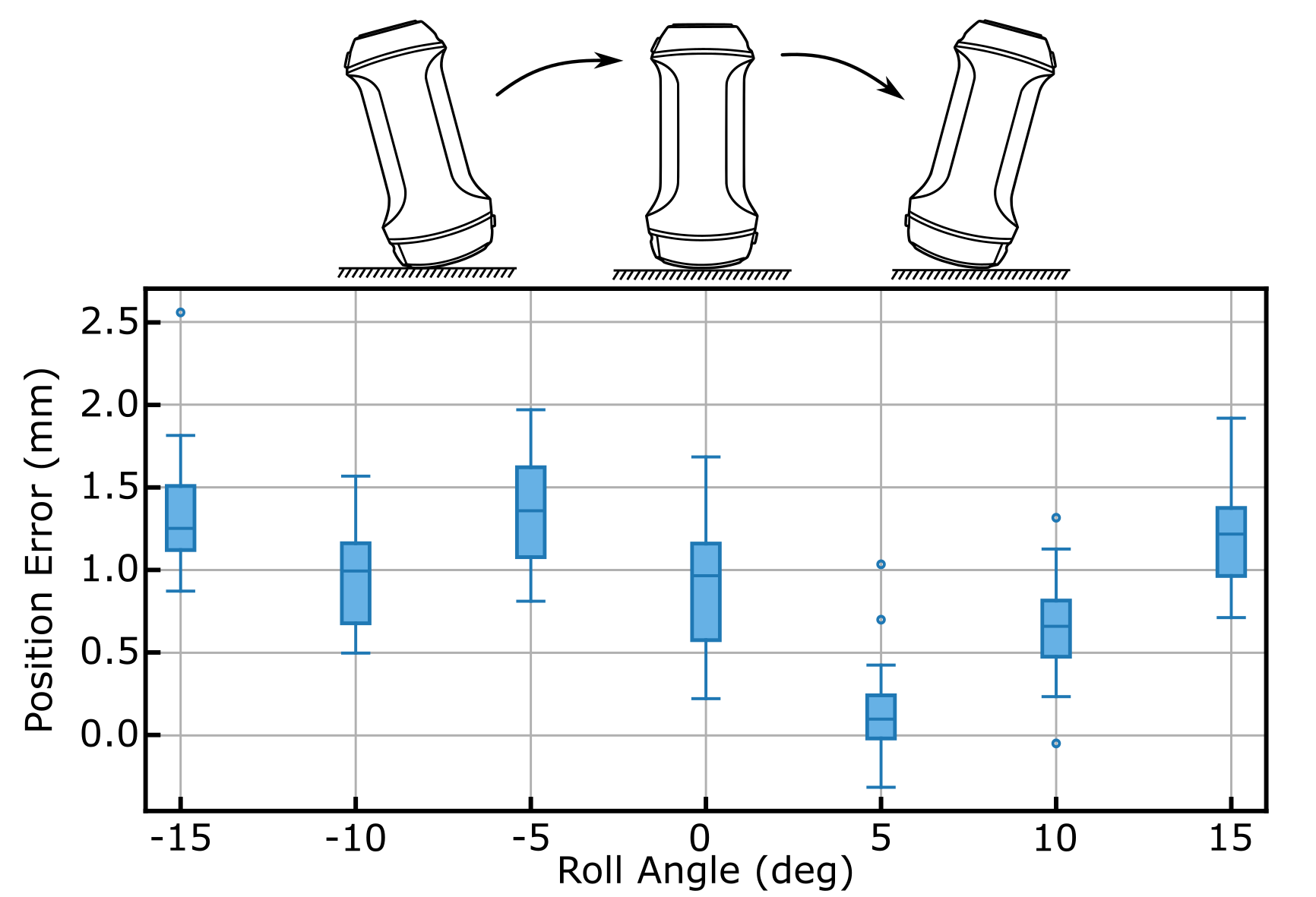}
\caption{Contact-point $Y$-position estimation error
across probe roll angles. }
\label{fig:cp_estimation_box}
\vspace{-10pt}
\end{figure}

Across five trials, the contact-point $Y$-position error relative to this geometric reference had an RMSE of $1.464 \pm 0.143$~mm and a mean signed error of $1.036 \pm 0.130$~mm, where the variation denotes the standard deviation across trials. The positive mean error indicates a systematic offset. Figure~\ref{fig:cp_estimation_box} shows the orientation-dependent error after averaging within $1^\circ$ roll-angle intervals. These results support the contact-point assumption under the tested conditions, with millimeter-level agreement between the image-derived estimate and the geometric reference.

\subsection{Static Controller Performance}

\begin{figure}[t]

\vspace*{5pt} 
\centering
\includegraphics[width=\linewidth]{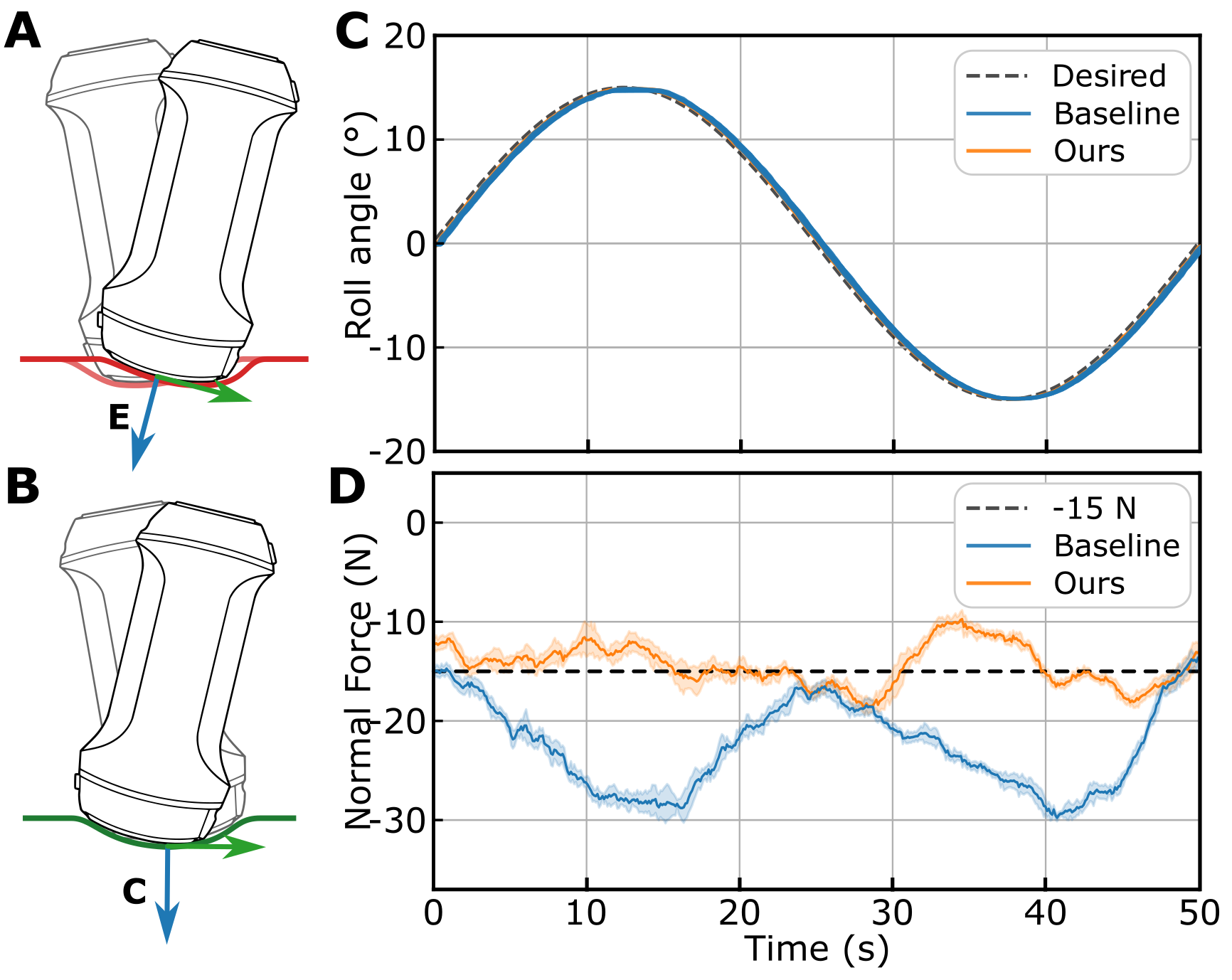}
\caption{Static controller performance comparison.
A) The baseline impedance controller regulates the fixed
end-effector tip point, producing additional penetration
during rolling (red curves).
B) The proposed controller updates the contact point along
the probe curvature (green curve).
C) Desired and measured probe roll trajectories.
D) Independently measured normal interaction forces.}
\label{fig:figure_6}
\vspace{-10pt}
\end{figure}

We compared the proposed controller with the baseline impedance
controller using the rolling protocol described in
Sec.~\ref{subsec:CPEstPer}. The baseline impedance controller
regulated the fixed end-effector tip point without adaptive
force taring. Rolling about this point introduces additional
penetration of the curved probe into the phantom, illustrated
by the red curves in Fig.~\ref{fig:figure_6}A. The proposed
controller instead updated the controlled contact point along
the probe's natural curvature, illustrated by the green curve
in Fig.~\ref{fig:figure_6}B.

The overall position error (RMSE ± raw-error SD) was \(1.115 \pm 0.457\) mm for the proposed controller and \(1.210 \pm 0.512\) mm for the baseline, while the roll error was \(0.360 \pm 0.360^\circ\) and \(0.369 \pm 0.368^\circ\), respectively. The position difference was statistically significant, whereas the roll difference was not (position: \(p<0.001\); roll: \(p=0.121\)). The roll trajectories are shown in Fig.~\ref{fig:figure_6}C.

The measured interaction forces differed substantially
(Fig.~\ref{fig:figure_6}D). For the baseline impedance controller,
compression increased with the magnitude of the roll angle
and returned toward the nominal level near $0^\circ$,
reaching a maximum of $31.56~\mathrm{N}$. The proposed
controller reduced this orientation dependence and limited
peak compression to $20.09~\mathrm{N}$, a $36.3\%$ reduction.
Relative to the nominal $15~\mathrm{N}$ magnitude, peak excess
compression decreased from $16.56$ to $5.09~\mathrm{N}$
($69.2\%$). These results demonstrate reduced roll-induced
loading with comparable tracking accuracy for the proposed
framework.

\subsection{Scanning Controller Performance}

We designed the scanning experiment to evaluate contact-point tracking during continuous scanning at different probe orientations and during roll transitions that shift the contact location along the curved probe surface. The probe first established contact with the phantom at a roll angle of $-15^\circ$ and performed a $50~\mathrm{mm}$ linear sweep at approximately $2.75~\mathrm{mm/s}$ while maintaining a constant nominal penetration depth. After each $50~\mathrm{mm}$ segment, the probe paused translation to change its roll angle, first to $0^\circ$
and then to $+15^\circ$, before continuing along the same scanning direction at the same speed. This sequence was performed separately along the $X$- and $Y$-axes. During roll transitions, the translational reference was held constant.

\begin{figure*}[t]
\vspace*{5pt} 
\centering
\includegraphics[width=\textwidth]{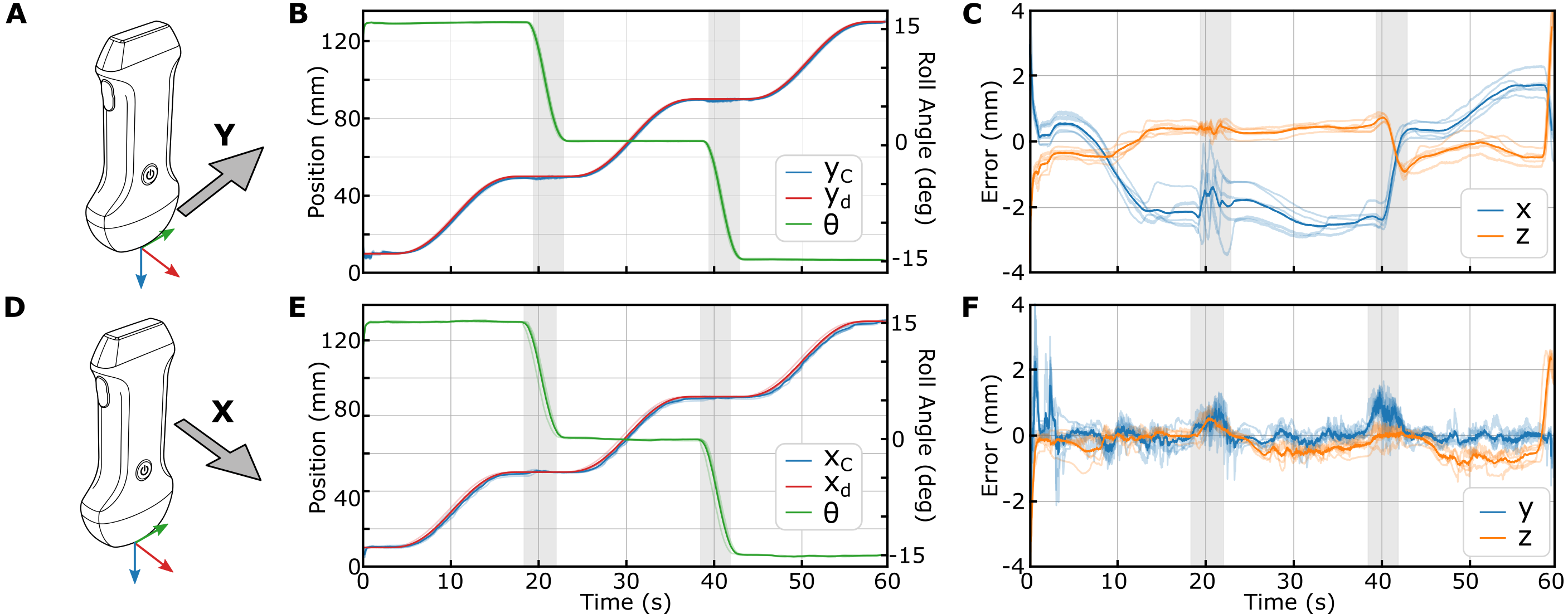}
\caption{Contact-point tracking during linear scanning
at different probe roll orientations.
(A) $Y$-directed scanning configuration.
(B) Desired and measured contact-point $Y$ position
and probe roll angle.
(C) Corresponding orthogonal $X$- and $Z$-axis
tracking errors.
(D) $X$-directed scanning configuration.
(E) Desired and measured contact-point $X$ position
and probe roll angle.
(F) Corresponding orthogonal $Y$- and $Z$-axis
tracking errors.
Shaded regions indicate the roll-transition intervals
during which the translational reference was held constant.}
\label{fig:sweeping}
\vspace{-10pt}
\end{figure*}

Our controller maintained millimeter-level contact-point
tracking during both scanning directions, as shown in
Fig.~\ref{fig:sweeping}. Across the five repeated trials,
the RMSE along the commanded scanning direction was
$0.93 \pm 0.08\,\mathrm{mm}$ for the $Y$-sweep and
$1.60 \pm 0.15\,\mathrm{mm}$ for the $X$-sweep.
During the $Y$-sweep, the orthogonal $X$-axis error was
larger, with an RMSE of $1.76 \pm 0.12\,\mathrm{mm}$,
whereas the corresponding $Y$-axis error during the
$X$-sweep was $0.25 \pm 0.01\,\mathrm{mm}$.
The surface-normal $Z$-axis RMSE remained below
$0.5\,\mathrm{mm}$ for both experiments, reaching
$0.40 \pm 0.05\,\mathrm{mm}$ and
$0.47 \pm 0.08\,\mathrm{mm}$ for the $Y$- and
$X$-sweeps, respectively.

As illustrated in Fig.~\ref{fig:sweeping}B and E,
the estimated contact point closely followed the commanded
scanning trajectory at each probe orientation. Larger
transient errors were primarily observed during the
intervening roll transitions, highlighted by the shaded
regions in Fig.~\ref{fig:sweeping}C and F, when the
effective contact location migrated along the curved
transducer surface while the translational reference
remained fixed. The maximum coordinate-wise contact-point
excursions during these transitions were
$(3.12,\,2.43,\,2.03)\,\mathrm{mm}$ along $(X,Y,Z)$
for the $Y$-sweep and
$(2.95,\,2.63,\,0.87)\,\mathrm{mm}$ for the $X$-sweep.
Following each reorientation, the tracking errors returned
toward their scanning levels as the subsequent linear
sweep was initiated.

These results demonstrate that the contact-aware controller
can regulate the image-derived contact point during continuous
scanning at different probe roll orientations, while limiting
out-of-plane tracking error and accommodating contact-location
migration during probe reorientation.

\section{Discussion}

This work uses B-mode ultrasound to provide both
contact-state feedback and a spatial reference for
robot--tissue interaction. The estimated contact location
updates the position regulated by the impedance controller,
while end-effector orientation is controlled independently.
This enables the controller to accommodate contact migration
along the curved probe during commanded reorientation.
Contact detection also provides an image-based event for
adaptive force taring, independently of the wrench estimate.

The rolling experiments illustrate why the choice of control
point matters: accurate tracking of a fixed end-effector point
alone does not prevent additional tissue loading when contact
moves along the probe surface. The proposed framework
maintained comparable tracking accuracy while reducing
peak compressive force by $36.3\%$. This improvement is
consistent with compensation for roll-induced geometric
displacement, although the comparison evaluates contact point updating and adaptive taring together. The scanning
experiments further demonstrated millimeter-level tracking
of the image-derived contact point during translation at different probe orientations, with larger transient errors during
roll transitions. Together, these findings support the use of
ultrasound images as spatial feedback for interaction control.

The current validation has several limitations. Performance
in heterogeneous tissue remains unverified: variations in
tissue contrast, acoustic shadowing, deformation, and imaging
parameters may affect both the contact-detection threshold
and the estimated contact location. Furthermore, the
non-contact phase relies on a stationary surface with known
geometry, limiting applicability when the surface is initially
unknown or changes during scanning. Curved surfaces also
introduce changes in the local surface normal and contact
geometry that require further evaluation.

Future work will address these limitations through richer
US image features and updated surface perception.
Deep learning-based image representations could improve
contact detection and localization across varying tissue
appearances, building on deep-feature-based ultrasound
visual servoing \cite{zakeri2025deep}.
Depth-camera-based surface perception could continuously
update the potential contact point during approach to
curved or moving surfaces \cite{zhetpissov2025asee2}.
Ultrasound elastography offers a complementary direction
for incorporating tissue deformation and mechanical
information into contact localization
\cite{groenhuis2020elastography}. Combining this information
with the acoustic contact estimate may improve prediction
of the mechanical center of pressure under heterogeneous
or asymmetric contact. These extensions will be evaluated
on curved and heterogeneous phantoms and, subsequently,
biological tissue.


\bibliographystyle{IEEEtran}
\bibliography{IEEEabrv,references}

\end{document}